\RequirePackage{amsmath}
\documentclass[runningheads]{llncs}
\usepackage[T1]{fontenc}
\usepackage{graphicx}
\usepackage{booktabs}
\usepackage{amssymb}
\usepackage{bm}
\usepackage{accsupp}
\usepackage[export]{adjustbox} 
\usepackage{multirow}
\usepackage{hyperref}
\usepackage{orcidlink}

\usepackage{color}
\usepackage{eso-pic}

\begin{document}
\AddToShipoutPictureBG*{%
  \AtPageLowerLeft{%
    \raisebox{1cm}{%
      \makebox[\paperwidth][c]{%
        \parbox{0.8\paperwidth}{\footnotesize\centering
          This is a post-peer-review, pre-copyedit version of an article
          published in CAIP 2025. The final authenticated version
          is available online at \url{https://doi.org/10.1007/978-3-032-04968-1_12}.
        }%
      }%
    }%
  }%
}

\title{RTFVE: Realtime Face Video Enhancement}
%
%
\author{Varun Ramesh Jois\orcidlink{0009-0004-2615-6406},
Antonella DiLillo,
James Storer}
\authorrunning{Jois et al.}
%
\institute{Brandeis University, Waltham MA 02453, USA
\email{\{vjois,dilant,storer\}@brandeis.edu}}
\maketitle              
\begin{abstract}
There's been a surge in adoption of video conferencing applications for both personal and business use cases. 
However, the bandwidth limitations faced by many users worldwide may restrict the optimal use of such applications.
Although deep learning offers a solution for enhancing low bit rate videos, most models today are either hard to incorporate with modern compression standards or require specialized hardware to run such as significant GPUs making these models impractical.
To address these issues, we introduce the Realtime Face Video Enhancement (RTFVE) model which can be easily incorporated with any video decoder and can run in realtime on ordinary CPUs. Experiments show that our model improves perceptual quality over the compressed video baseline at multiple low bitrate settings. 
The source code will be made available at \url{https://github.com/varun-jois/RTFVE}.

\end{abstract}
\section{Introduction}
\label{sec:intro}
Over the last decade videocalling has become one of the most popular forms of communication. 
With today's technology, everyone has access to a phone with a front-facing camera and an internet connection, often making video calling a daily occurrence for many users. Ever since the lockdowns during the COVID-19 pandemic, the adoption of this technology has grown exponentially with many new companies providing services for it. 

While there has been a global trend towards faster internet speeds, there are still millions of people around the world that experience slow internet connections. And while the video compression standards are flexible with handling low bandwidth settings, they come with the cost of low quality video quite often making videocalls untenable. This being the case even though many of these users have the basic hardware to handle a standard videocall. 

Deep learning based methods hold promise when it comes to improving the perceptual quality of faces. There have been advances in a myriad of fields such as face restoration, face deblurring, face super-resolution, etc. However, when it comes to face enhancement for videocalls there have been two problems stymieing the adoption of deep learning: 1) Models that cannot be easily integrated into the ubiquitous compression standards and 2) Models that require significant GPU and NPU resources that typical users don't have. In this paper, we address both of these issues.

Our contributions are as follows:
\begin{enumerate}
    \item We present our model \textbf{R}eal\textbf{T}ime \textbf{F}ace \textbf{V}ideo \textbf{E}nhance-ment (RTFVE) a model for face enhancement that can easily be integrated with the decoder of a video compression standard.
    \item For the bandwidth cost of a handful of high-quality reference frames, our model is able to improve the visual quality of faces over the compression standards running in realtime on typical CPUs (e.g. laptops).
\end{enumerate}

\begin{table}[tb]
    \caption{Relative speed comparison of various methods ranging from classical computer vision algorithms to current state-of-the-art face restoration models on a low cost CPU. For realtime performance an FPS of at least 24 is required.}
    \label{tab:fps}
    \centering
    \begin{tabular}{|c|c|c|}
        \hline
        \textbf{Model} & \textbf{Type} & \textbf{Frames per Second (↑)} \\
        \hline
        Non-Local Means Denoising~\cite{nlmd-slow} & Classical & 4.56 \\
        \hline
        Bilateral Filtering~\cite{bilateral} & Classical & 13.89 \\
        \hline
        Split-Bregman~\cite{bregman} & Classical & 17.71 \\
        \hline
        Chambolle~\cite{Chambolle2004} & Classical & 7.60 \\
        \hline
        Codeformer~\cite{codeformer} & Neural Network & 0.05 \\
        \hline
        GPEN~\cite{gpen} & Neural Network & 2.47 \\
        \hline
        GFP-GAN~\cite{gfpgan} & Neural Network & 2.02 \\
        \hline
        \textbf{RTFVE (Ours)} & Neural Network & 24.85 \\
        \hline
    \end{tabular}
\end{table}

\section{Related Work}
\label{sec:format}
\subsection{Face Restoration}
Face restoration is the task of producing a high quality image of a face from its low quality counterpart. 
GPEN~\cite{gpen} works by embedding a trained GAN-prior-network as the decoder of a U-shaped DNN and then fine-tuning the DNN with synthesized low quality face images. 
CodeFormer~\cite{codeformer} looks at the blind face restoration task as a codebook prediction task. They first learn a discrete codebook and decoder to store high quality parts of faces. With the codebook and decoder fixed, they train a transformer based model to predict the code sequence of the low quality input image thereby modeling the properties of the low quality faces. PGDiff~\cite{pgdiff} introduces the concept of partial guidance to the diffusion process by modeling properties such as face structure and color statistics and applying this guidance during the reverse diffusion process.
FRR-Net~\cite{frrnet} is a face restoration and lighting network that incorporates a distortion classifier to use the class as a prior and dice loss for segmentation masks to only focus on the face region. They propose a new degradation scheme that also include illumination distortions.

\subsection{Video Restoration}
MFQE~\cite{mfqe} leverages a BiLSTM based detector to locate Peak Quality Frames (PQFs) in compressed video and then uses a Multi-Frame Convolutional Neural Network to enhance the quality of compressed video using neighboring PQFs. They use multi scale features and dense connections to improve enhancement performance. In EDVR~\cite{edvr} the authors perform feature alignment using a pyramid, cascading, deformable convolution block (PCD). They then combine the aligned frame features using the temporal and spatial attention module (TSA) to perform attention both temporally and spatially before finally reconstructing the output. BasicVSR~\cite{basicvsr} uses bidirectional propagation, flow-based alignment, concatenation based aggregation and pixel shuffle to perform restoration. \cite{Agnolucci} uses keyframes as references and computes multiscale features which are used to produce the final output.

\begin{figure*}[tb]
    \centering
    \BeginAccSupp{ActualText={The two branches of the model. One branch produces features for the high quality reference images. These features are cached and repeatedly sent to the main branch of the model that also takes as input the low quality frame and produces the enhanced output.}}
    \includegraphics[width=\linewidth]{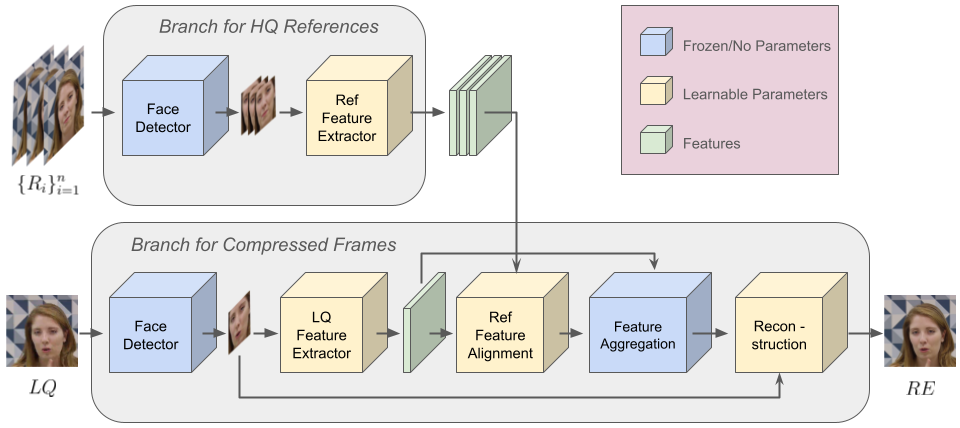}
    \EndAccSupp{}
    \caption{The Realtime Face Video Enhancement (RTFVE) model. Our model takes the compressed, low-quality frames $LQ$ from a video call, and $n$ high-quality reference frames $\{R_i\}_{i=1}^{n}$ to produce the enhanced reconstructed frame $RE$.}
    \label{fig:framework}
\end{figure*}

\section{RTFVE Framework}
\label{sec:pagestyle}
Our model is inspired by previous work that leverages reference images such as \cite{C2,jois} for Super-Resolution and \cite{Agnolucci} for Face Video Enhancement. 
The objective of our model is to leverage a handful of high quality reference images to improve the aesthetic of a low quality video stream. An equally important objective of ours is to make our model practical in terms of latency and hardware requirements. 
A model that is not at least 24 frames per second may diminish
the user experience and a model that requires substantial GPU power to run may not be practical for the limited resources of many users.

Our model, named Realtime Face Video Enhancement (RTFVE) is depicted in Figure~\ref{fig:framework}. It is comprised of five parts; a face detector, feature extraction blocks, an alignment module, a feature aggregation module and a reconstruction module. The model takes a low quality image $LQ$ and $n$ high quality reference images $\{R_i\}_{i=1}^{n}$ as inputs to produce the enhanced reconstructed frame $RE$.

\subsection{Face Detector}
\label{face detector}
The first part of our model is the face detector $f_{d}$ which locates and crops the face in the low quality and reference images and passes them along to the rest of the model. 
\begin{equation}
\begin{split}
    LQ^\text{face} &= f_{d}(LQ) \\
    R^\text{face}_i &= f_{d}(R_i) \quad \textrm{for} \quad i=1...n
\end{split}
\end{equation}
By working on just the faces, we can considerably reduce the resolution of our inputs thereby reducing latency. In our model, we select a $256\times256$ size crop. For our work, we used an off the shelf face detector \cite{blaze} that we fixed. We also add a gaussian blur to the low quality image using a kernel of shape $3\times3$ and $\sigma=0.8$.

\subsection{Feature Extractor and the Shuffle Unit}
After obtaining the face crops, we first perform pixel unshuffling~\cite{pix_shuffle} to reduce the resolution before passing them to the feature extractors $f_e$. This is the reverse of the shuffling operation mentioned in~\cite{pix_shuffle} that is a popular method for reducing the resolution size without losing any data by increasing the number of channels. By working in lower resolution space, we greatly increase speed of the model. Here we reduce the resolution by $4\times$. We use different extractors for the LQ and reference images. 
\begin{equation}
\begin{split}
    F_{LQ} &= f^{LQ}_{e}(LQ^\text{face}) \\
    F_{R_i} &= f^{R}_{e}(R^\text{face}_i) \quad \textrm{for} \quad i=1...n
\end{split}
\end{equation}
Feature extraction is a necessary step for neural network architectures and deeper features offer better semantic representation. But deep models are computationally expensive making it unfeasible for low latency. To overcome this we designed the extractors, as well as the other modules using shuffle units~\cite{shufflenetv2}. These units are analogous to the units in ResNet~\cite{resnet} with three convolutional layers per unit. However, they greatly differ in the number of parameters and speed since the former performs depthwise group convolutions and $1\times1$ convolutions instead of dense convolutions. The shuffle unit also performs a channel split operation working on half the number of channels in any block thereby further increasing speed. For our features extractors, we used 10 shuffle units giving us deeper features and speed.

\subsection{Feature Alignment}
To get the most information out of the reference features, they need to be aligned with the low quality image features. We perform the alignment using spatial transformer alignment \cite{jois} $f_a$:
\begin{equation}
    F_{R_iA} = f_a(F_{LQ}, F_{R_i}) \quad \textrm{for} \quad i=1...n
\end{equation}
This module works by aligning the features of the high quality reference images $\{F_{R_i}\}_{i=1}^{n}$ with the features of the low quality frame $F_{LQ}$ in feature space. The original alignment module had the major drawback of two fully connected layers. This drastically increased the number of parameters and computation time. To remedy this, we used downsampling shuffle units within the localization network so the number of parameters given to the fully connected layers is low.

\begin{table}[h]
\caption{Performance scores with respect to video compression codec and compression rate. Higher the CRF, more the compression.}
\label{tab:scores}
\centering
\begin{tabular}{|c|c|c|c|c|c|}
\hline
\textbf{Codec} & \textbf{CRF} & \textbf{Model} & \textbf{PSNR (↑)} & \textbf{SSIM (↑)} & \textbf{LPIPS (↓)} \\
\hline\hline
\multirow{9}{*}{H.264} & \multirow{3}{*}{36} & H.264 (baseline) & 33.0196 & 0.9089 & 0.0826 \\
 & & RTFVE & 33.5172 & 0.92 & 0.1073 \\
 & & RTFVE-gan & 33.3745 & 0.9118 & 0.0715 \\
\cline{2-6}
 & \multirow{3}{*}{40} & H.264 (baseline) & 30.833 & 0.8752 & 0.1273 \\
 & & RTFVE & 31.6259 & 0.8961 & 0.1365 \\
 & & RTFVE-gan & 31.2276 & 0.8799 & 0.1076 \\
\cline{2-6}
 & \multirow{3}{*}{44} & H.264 (baseline) & 28.566 & 0.8307 & 0.1872 \\
 & & RTFVE & 29.3205 & 0.8586 & 0.1851 \\
 & & RTFVE-gan & 29.0675 & 0.8433 & 0.1447 \\
\hline\hline
\multirow{9}{*}{H.265} & \multirow{3}{*}{36} & H.265 (baseline) & 33.0416 & 0.9131 & 0.0779 \\
 & & RTFVE & 33.7089 & 0.9271 & 0.0926 \\
 & & RTFVE-gan & 33.1195 & 0.9086 & 0.0899 \\
\cline{2-6}
 & \multirow{3}{*}{40} & H.265 (baseline) & 30.7886 & 0.8788 & 0.1226 \\
 & & RTFVE & 31.5147 & 0.8993 & 0.133 \\
 & & RTFVE-gan & 31.0642 & 0.8813 & 0.1106 \\
\cline{2-6}
 & \multirow{3}{*}{44} & H.265 (baseline) & 28.5467 & 0.8353 & 0.1899 \\
 & & RTFVE & 29.2606 & 0.8631 & 0.1876 \\
 & & RTFVE-gan & 28.9606 & 0.8468 & 0.1434 \\
\hline
\end{tabular}
\end{table}

\subsection{Feature Aggregation}
With the aligned reference features $\{F_{R_iA}\}_{i=1}^{n}$ in hand, we can now aggregate our references to obtain the most useful signals. We adopt the weighted aggregation module $f_{agg}$ in \cite{jois} as is: since it's parameter free and efficient. 
\begin{equation}
    F_{agg} = f_{agg}(F_{LQ}, \{F_{R_iA}\}_{i=1}^{n}) 
\end{equation}
This module works by giving a greater weight to the reference features that are closest (in terms of L2 distance) to the features of the low-quality frame. This weighting is done for all regions in feature space and so one feature map can be given a greater weight at one region whereas another feature map can be given a greater weight in another region. It should be noted that this module does not contain parameters as it is based on the L2 distance and the softmax function and can be executed in a few steps efficiently.

\subsection{Reconstruction Module}
We pass the aggregated features $F_{agg}$ to the reconstruction module $f_r$ which produces the residuals that get added to the input image. To match the resolutions, a pixel shuffle~\cite{pix_shuffle} is produced at the end of the reconstruction block.
\begin{equation}
    RE^\text{face} = LQ^\text{face} + f_r({F_{agg}})
\end{equation}
Similar to other papers \cite{Agnolucci,jois} we find producing residuals with the model more effective than directly producing the output. The reconstruction module is composed of 10 shuffle units. Finally, we replace the low-quality face crop with the the enhanced face crop $RE^\text{face}$ to produce the final output $RE$.

\subsection{Enhanced Speed For Subsequent Frames}
One integral aspect of the model is the ability to reuse reference features for subsequent frames. Once features have been extracted from the high quality reference images, they can be cached and used throughout the video without having to go through the feature extractor. With these reference features, we can bypass the detector and extraction steps and directly move to the alignment step saving computation over the course of a call.

\begin{figure*}[t]
    \centering
    \BeginAccSupp{ActualText={Frames from a video call. The first row contains the low quality frames from the compressed video. The videos were compressed with a Constant Rate Factor of 40. The second row contains the enhanced versions of the low quality frames produced by our model. The third and final row contains the ground truth frames. The model is able to considerably improve the quality of the input video.}}
    \parbox[b]{0.15\linewidth}{\raggedright Low Quality Frame}\hfil
    \parbox[b]{0.15\linewidth}{\centering\includegraphics[width=\linewidth, valign=c]{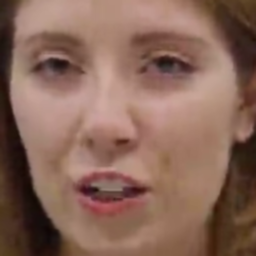}}\hfil
    \parbox[b]{0.15\linewidth}{\centering\includegraphics[width=\linewidth, valign=c]{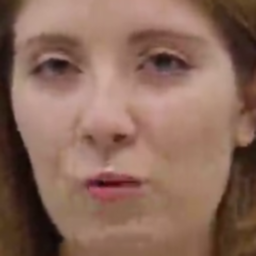}}\hfil
    \parbox[b]{0.15\linewidth}{\centering\includegraphics[width=\linewidth, valign=c]{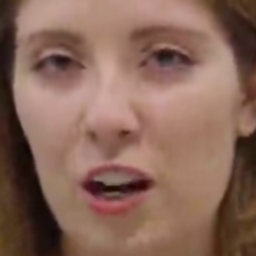}}\hfil
    \parbox[b]{0.15\linewidth}{\centering\includegraphics[width=\linewidth, valign=c]{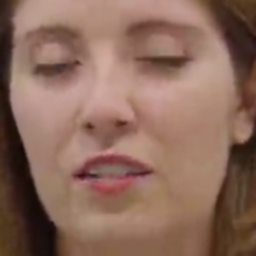}}\hfil
    \parbox[b]{0.15\linewidth}{\centering\includegraphics[width=\linewidth, valign=c]{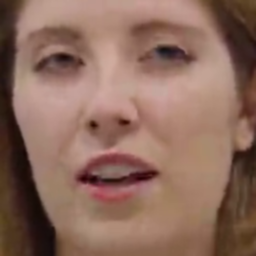}}\\[1ex]
    
    \parbox[b]{0.15\linewidth}{\raggedright Enhanced Frame}\hfil
    \parbox[b]{0.15\linewidth}{\centering\includegraphics[width=\linewidth, valign=c]{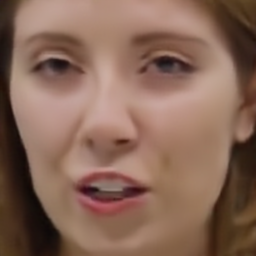}}\hfil
    \parbox[b]{0.15\linewidth}{\centering\includegraphics[width=\linewidth, valign=c]{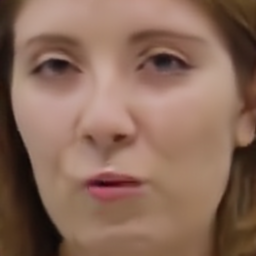}}\hfil
    \parbox[b]{0.15\linewidth}{\centering\includegraphics[width=\linewidth, valign=c]{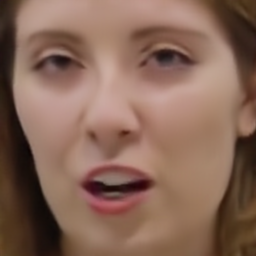}}\hfil
    \parbox[b]{0.15\linewidth}{\centering\includegraphics[width=\linewidth, valign=c]{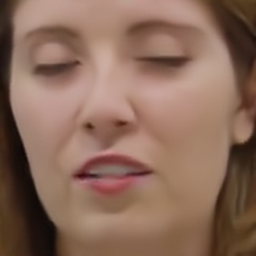}}\hfil
    \parbox[b]{0.15\linewidth}{\centering\includegraphics[width=\linewidth, valign=c]{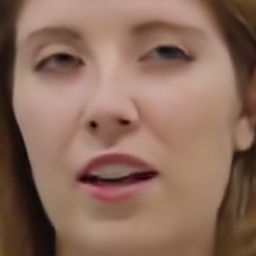}}\\[1ex]
    
    \parbox[b]{0.15\linewidth}{\raggedright Ground Truth}\hfil
    \parbox[b]{0.15\linewidth}{\centering\includegraphics[width=\linewidth, valign=c]{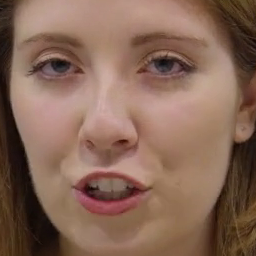}}\hfil
    \parbox[b]{0.15\linewidth}{\centering\includegraphics[width=\linewidth, valign=c]{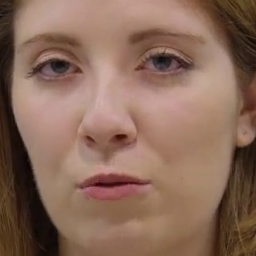}}\hfil
    \parbox[b]{0.15\linewidth}{\centering\includegraphics[width=\linewidth, valign=c]{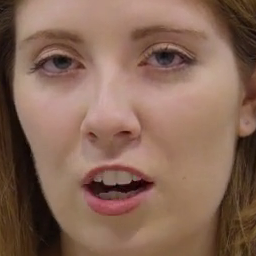}}\hfil
    \parbox[b]{0.15\linewidth}{\centering\includegraphics[width=\linewidth, valign=c]{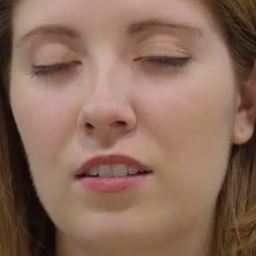}}\hfil
    \parbox[b]{0.15\linewidth}{\centering\includegraphics[width=\linewidth, valign=c]{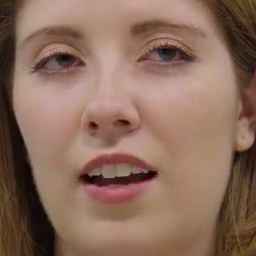}}

    \EndAccSupp{}
    \caption{Stills from videos compressed at CRF 40.}
    \label{fig:crf40}
\end{figure*}

\section{Experiments}
\label{sec:EXPERIMENTD}
\subsection{Datasets}
Similar to \cite{Agnolucci} and \cite{jois}, we use the publicly available DeepFakeDetection dataset \cite{DFDdataset} to perform our experiments. The dataset is comprised of 363 1080p videos that were made using 28 paid actors. Out of these videos, we used the videos belonging to the categories \textit{"outside talking still laughing"}, \textit{"podium speech happy"} and \textit{"talking against wall"} as these were the ones most similar to a video call setting. The first 22 identities corresponding to 65 videos were used as the training set and the remaining 6 identities corresponding to 17 videos were used as the testing set.

For all videos, we first cropped a $512\times512$ pixel region around the human subject, extracted 5 frames from the first 12 seconds of the video at a sampling rate of 72 frames (input videos have a frame rate of 24 frames per second) to form our reference images, skipped 3 seconds and then compressed the rest of the video. We chose this policy as it most resembles a real videocall scenario where the high quality frames can be sent early in the stream, with the rest of the stream being compressed. While we extracted 5 reference frames, we only used 3 in our model. Some frames didn't contain the entire face so we extracted 2 more to have more options to choose from.

Our low quality videos were compressed at three different compression rates using two of the most popular video compression standards - H.264 and H.265. We used ffmpeg with the libx264 encoder for H.264 and libx265 for H.265. The compression rates we chose were a Constant Rate Factor (CRF) of 36, 40 and 44 to mimic a range of low bandwidth settings. For reference, the CRF parameter ranges from 0-51 where 0 is lossless, 18 is considered visually lossless and 51 is the worst quality possible. Generally speaking, a CRF of 17-28 is considered an acceptable range according to the ffmpeg video encoding guide. The reason we chose to experiment with H.264 is because it is still a widely adopted codec for videocalls with the greatest support across a range of devices including older Android and iOS phones. H.264 and its variants are also the codecs used by popular videoconferencing applications such as Zoom and Webex. By including H.265 video in our tests, we demonstrate that our model operates reliably irrespective of the codec used.

To speed up training, instead of training on the entire video, we trained on 20 frames from each video sampled at a rate of 10 frames. We also trained our model on a tighter cropped region around the face with a resolution of $256\times256$. These were obtained using the aforementioned face detector~\ref{face detector}.

\begin{figure*}[t]
    \centering
    \BeginAccSupp{ActualText={Frames from a video call. The first row contains the low quality frames from the compressed video. The videos were compressed with a Constant Rate Factor of 44. The second row contains the enhanced versions of the low quality frames produced by our model. The third and final row contains the ground truth frames. The model is able to considerably improve the quality of the input video despite the high compression.}}
    \parbox[b]{0.15\linewidth}{\raggedright Low Quality Frame}\hfil
    \parbox[b]{0.15\linewidth}{\centering\includegraphics[width=\linewidth, valign=c]{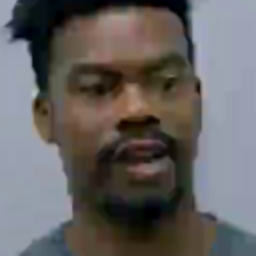}}\hfil
    \parbox[b]{0.15\linewidth}{\centering\includegraphics[width=\linewidth, valign=c]{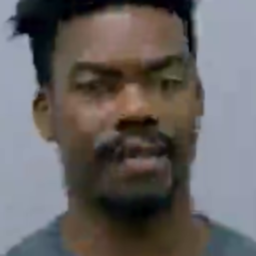}}\hfil
    \parbox[b]{0.15\linewidth}{\centering\includegraphics[width=\linewidth, valign=c]{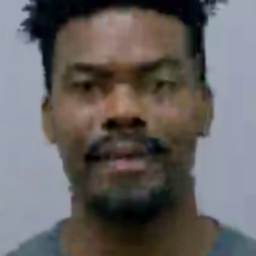}}\hfil
    \parbox[b]{0.15\linewidth}{\centering\includegraphics[width=\linewidth, valign=c]{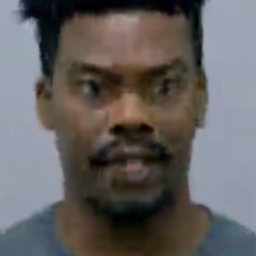}}\hfil
    \parbox[b]{0.15\linewidth}{\centering\includegraphics[width=\linewidth, valign=c]{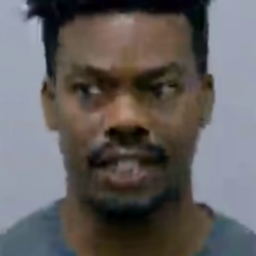}}\\[1ex]
    
    \parbox[b]{0.15\linewidth}{\raggedright Enhanced Frame}\hfil
    \parbox[b]{0.15\linewidth}{\centering\includegraphics[width=\linewidth, valign=c]{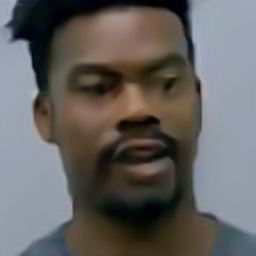}}\hfil
    \parbox[b]{0.15\linewidth}{\centering\includegraphics[width=\linewidth, valign=c]{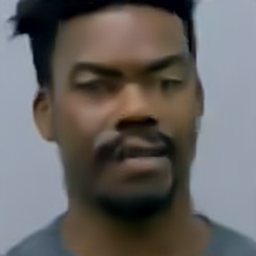}}\hfil
    \parbox[b]{0.15\linewidth}{\centering\includegraphics[width=\linewidth, valign=c]{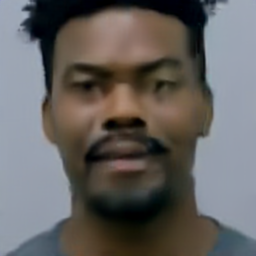}}\hfil
    \parbox[b]{0.15\linewidth}{\centering\includegraphics[width=\linewidth, valign=c]{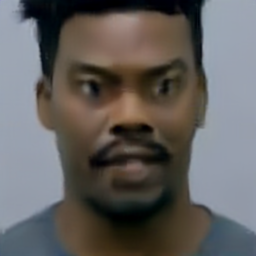}}\hfil
    \parbox[b]{0.15\linewidth}{\centering\includegraphics[width=\linewidth, valign=c]{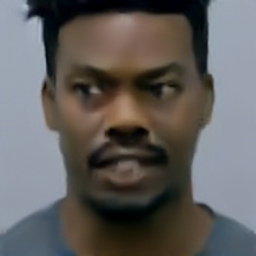}}\\[1ex]
    
    \parbox[b]{0.15\linewidth}{\raggedright Ground Truth}\hfil
    \parbox[b]{0.15\linewidth}{\centering\includegraphics[width=\linewidth, valign=c]{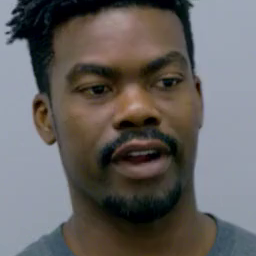}}\hfil
    \parbox[b]{0.15\linewidth}{\centering\includegraphics[width=\linewidth, valign=c]{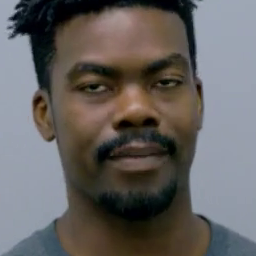}}\hfil
    \parbox[b]{0.15\linewidth}{\centering\includegraphics[width=\linewidth, valign=c]{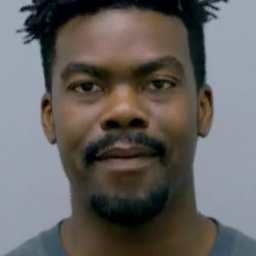}}\hfil
    \parbox[b]{0.15\linewidth}{\centering\includegraphics[width=\linewidth, valign=c]{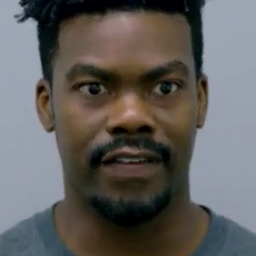}}\hfil
    \parbox[b]{0.15\linewidth}{\centering\includegraphics[width=\linewidth, valign=c]{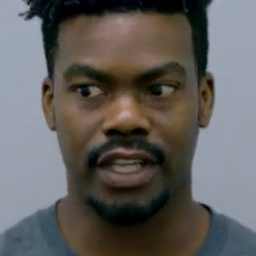}}

    \EndAccSupp{}
    \caption{Stills from videos compressed at CRF 44.}
    \label{fig:crf44}
\end{figure*}

\subsection{Training}
We trained our model for 1000 epochs using the Adam optimizer with a learning rate of $10^{-4}$ and the default $\beta_1=0.9$, $\beta_2=0.999$. To improve the performance of our model, we performed a number of random augmentations including, random horizontal flip, random vertical flip, and a random affine transformation with a rotation in the range [-5\textdegree, 5\textdegree], translation along the horizontal and vertical axes in the range [-2\%, 2\%] and scaling in the range of [0.95, 1.05]. No shearing was performed. 
Our model can take any number 1 or more reference images; for our
experiments, we use 3 reference images.

We compared our model's performance to the baseline input compressed video. While there are numerous works on face restoration and face enhancement, to the best of our knowledge, none of these works can run in realtime (at least 24 FPS) on a typical CPU, and thus we didn’t evaluate comparisons against them. In this work we are primarily concerned with face enhancement in constrained resource settings.

The recreation loss $l_{rec}$ we optimized for was the Mean Absolute Error which is given by:
\begin{equation}
    l_{rec} = \frac{1}{HWC}\left \lVert HQ - RE \right \rVert_1
\end{equation}
where $HQ$ is the high quality ground truth frame and $RE$ is the reconstructed frame. $HWC$ is the height, width and number of channels in $HQ$ and $RE$. We also performed adversarial training on our model based on the LS-GAN \cite{lsgan} where discriminator loss $l_{dis}$ and generator loss $l_{gen}$ are given by:

\begin{equation}
\begin{split}
    l_{dis} = \frac{1}{2}\mathbb{E}_{\bm{HQ} \sim \bm{P}(\bm{HQ})}\bigl[(D(HQ)-1)^2\bigr] + \\
    \frac{1}{2}\mathbb{E}_{\bm{LQ} \sim \bm{P}(\bm{LQ})}\bigl[D(G(LQ))^2\bigr]
\end{split}
\end{equation}

\begin{equation}
    l_{gen} = \frac{1}{2}\mathbb{E}_{\bm{LQ} \sim \bm{P}(\bm{LQ})}\bigl[(D(G(LQ))-1)^2\bigr]
\end{equation}

The total loss $l_{total}$ for our GAN based model is given by:
\begin{equation}
    l_{total} = l_{rec} + 0.01 * l_{gen}
\end{equation}

\subsection{Quantitative Results}
To evaluate our models we considered both distortion and perceptual metrics. For distortion, we measured Peak Signal to Noise Ratio (PSNR), Structural Similarity (SSIM) and for a perceptual metric, we measured Learned Perceptual Image Patch Similarity (LPIPS)~\cite{lpips}. We used the version of LPIPS that uses the AlexNet~\cite{alex} backbone. The scores are reported in table \ref{tab:scores}. All scores reported on a $256\times256$ face region. Our RTFVE model, when trained solely with reconstruction loss, achieves PSNR improvements of approximately 0.5, 0.8, and 0.76 dB for CRF values 36, 40, and 44 on the H.264 datasets respectively. Similarly, for videos compressed with the H.265 standard, the model yields PSNR gains of about 0.66, 0.73, and 0.72 dB for CRF 36, 40, and 44 respectively. These results confirm that our model delivers consistent performance improvements regardless of the codec used. Similarly, there is also an increase in SSIM. This is impressive given the severity of degradation found in the CRF 40 and 44 videos. Our GAN model, named RTFVE-gan, is not only able to increase PSNR and SSIM but is also able to reduce LPIPS scores which is considered better for this metric. We found that when the model has fewer parameters to train, it becomes harder to lower LPIPS scores so given the low number of parameters in our model, lowering LPIPS is also impressive.

Table \ref{tab:fps} shows the speed of our model on CPU compared to other state-of-the-art face restoration models. Our frame rates include the time taken by the face detector model to find the face. With minimal inference optimizations and no quantization our model runs at a median of 24.85 frames per second on a circa 2018 laptop with an Intel® Core™ i7-8550U CPU @ 1.80GHz × 8. It is reasonable to assume that a typical video call user may have this level of resource or better, and thus in practice our model runs in realtime and is suited for most everyday settings where a GPU or NPU is unavailable. On the other hand, typical face restoration models are not designed for low latency and low compute environments and hence fall well short of the 24 frames per second required for realtime applications. Our model also consumes low memory and is only 190K parameters - less than a megabyte of disk space. When reference features are recalculated for each frame, speed falls to 19.04 fps, which shows how our caching design improves inference speed.

Since our model, to the best of our knowledge, is the only model capable of running in realtime on CPU we couldn't compare our model with other face restoration models. Even classic, handcrafted image processing algorithms such as Non-Local Means Denoising~\cite{nlmd-slow}, Bilateral Filtering~\cite{bilateral} and total variation denoising methods such as Split Bregman~\cite{bregman} and Chambolle~\cite{Chambolle2004} do not run in realtime~\ref{tab:fps}. We did however compare against low-pass filters that have lower complexity and can be run in realtime. For this experiment we considered H.264 videos compressed at a CRF of 40. As we can see in table~\ref{tab:lowpass}, many of these filters produce worse results compared to the given compressed video (baseline). Only the Gaussian Filter was able to marginally improve performance and is well short of our model's output.

\begin{table}[tb]
\caption{Various realtime methods on H.264 CRF 40 data.}
\label{tab:lowpass}
\centering
\begin{tabular}{|c|c|c|c|}
\hline
\textbf{Model} & \textbf{PSNR (↑)} & \textbf{SSIM (↑)} & \textbf{LPIPS (↓)} \\
\hline
H.264 (baseline) & 30.833 & 0.8752 & 0.1273 \\
\hline
Mean Filter & 28.2071 & 0.8787 & 0.2259 \\
\hline
Median Filter & 30.5624 & 0.8766 & 0.1519 \\
\hline
Wavelet Denoising & 30.848 & 0.8762 & 0.1288 \\
\hline
Gaussian Filter & 30.964 & 0.8821 & 0.1623 \\
\hline
\textbf{RTFVE (Ours)} & 31.6259 & 0.8961 & 0.1365 \\
\hline
\end{tabular}
\end{table}

\subsection{Qualitative Results}
Figures \ref{fig:crf40} and \ref{fig:crf44} compare quality between the compressed and enhanced frames. These stills were taken from videos compressed using the H.264 standard. The reader is encouraged to view these figures realistically enlarged on a screen. 
Figure \ref{fig:crf40} shows the output stills of our model for a video compressed at CRF 40. From the first row, we can see that the input low quality frame has many artifacts including unnatural skin texture and blocking artifacts around the mouth. In the second row we can see that the model was able to alleviate these issues by producing smoother looking skin and has considerably reduced the blocking around the mouth, nose and chin. The output produced is also considerably sharper with the individual parts of the face being more well defined. In figure \ref{fig:crf44} we can see the model's performance on a CRF 44 video. Here the input is heavily deformed with large blocking artifacts and speckly skin texture. The model again does a good job smoothing the skin and improving the definition of the face under these difficult conditions. The improved quality is arguably more significant than the improvement in PSNR suggests. A video demo is available at \url{https://sigport.org/documents/rtfve}.

\section{Conclusion}
This paper introduces a novel lightweight model for videocall face enhancement. Unlike previous models for face enhancement, our model can run in realtime on typical user devices (e.g. laptop CPU) making it widely applicable. Our model is able to objectively and subjectively improve the video call experience under constrained circumstances where compression rates are high and visual quality is low. It achieves this by leveraging the information contained in high quality reference frames which are very similar to the low quality frames in the input stream. Our model can be seamlessly integrated with current compression standards where the model would be taking as input the output of the decoder.

For future work, we would like to find ways of further improving quality under constrained circumstances. Obtaining better perceptual quality with small models is challenging due to a lack of modeling ability but we seek to find ways to better exploit high quality reference images. We are also interested in building realtime models for larger resolutions.

\begin{credits}
\subsubsection{\discintname}
The authors have no competing interests to declare that are
relevant to the content of this article.
\end{credits}
%
%
%
\bibliographystyle{splncs04}
\bibliography{main}
\end{document}